\documentclass[conference]{IEEEtran}

\IEEEoverridecommandlockouts
\usepackage{cite}

\usepackage{amsmath,amssymb,amsfonts}
\usepackage{algorithmic}

\usepackage{subfig} 

\usepackage{graphicx}
\usepackage{textcomp}
\usepackage{xcolor}
\def\BibTeX{{\rm B\kern-.05em{\sc i\kern-.025em b}\kern-.08em
    T\kern-.1667em\lower.7ex\hbox{E}\kern-.125emX}}
\begin{document}

\title{Robustness in Compressed Neural Networks for Object Detection

\thanks{This work has been partially supported by Statutory Funds of Electronics, Telecommunications and Informatics Faculty, Gdańsk University of Technology. This work was supported in part by the Polish National Centre for Research and Development (NCBR) through the European Regional Development Fund entitled: INFOLIGHT Cloud-Based Lighting System for Smart Cities under Grant POIR.04.01.04/2019.}
}

\author{\IEEEauthorblockN{Sebastian Cygert}
\IEEEauthorblockA{\textit{Multimedia Systems Department} \\
Gdańsk, Poland \\
sebcyg@multimed.org}
\and
\IEEEauthorblockN{Andrzej Czyżewski}
\IEEEauthorblockA{\textit{Multimedia Systems Department} \\
\textit{Gdańsk University of Technology}\\}
}


\maketitle

\begin{abstract}
 Model compression techniques allow to significantly reduce the computational cost associated with data processing by deep neural networks with only a minor decrease in average accuracy. Simultaneously, reducing the model size may have a large effect on noisy cases or objects belonging to less frequent classes. It is a crucial problem from the perspective of the models’ safety, especially for object detection in the autonomous driving setting, which is considered in this work. 

It was shown in the paper that the sensitivity of compressed models to different distortion types is nuanced, and some of the corruptions are heavily impacted by the compression methods (i.e., additive noise), while others (blur effect) are only slightly affected. A common way to improve the robustness of models is to use data augmentation, which was confirmed to positively affect models’ robustness, also for highly compressed models. It was further shown that while data imbalance methods brought only a slight increase in accuracy for the baseline model (without compression), the impact was more striking at higher compression rates for the structured pruning. Finally, methods for handling data imbalance brought a significant improvement of the pruned models' worst-detected class accuracy.   

\end{abstract}


\section{Introduction}

Optimization of the size of visual recognition models is of great importance, for example, for autonomous driving, because of energy consumption, hardware cost, and size. Typical methods to reduce the computation cost include deploying specialized architectures \cite{mobilenets}, model compression techniques such as reducing model precision (quantisation) \cite{quantization} and/or setting the number of weight or filters to zero (pruning) \cite{braindamage}. For example Han et. al showed \cite{TODOnipsprune}, that it is possible to reduce the size of the VGG network by a factor of 13 when benchmarking on the ImageNet dataset \cite{imagenet} with no loss in accuracy.

Another essential aspect for real-world deployment is the models’ robustness. Many works have shown that current machine learning models for visual recognition from RGB images are vulnerable to tiny changes in the input image, such as adversarial examples  \cite{Zaremba}, noisy input \cite{Corruptions, fourier} small transformations of the input image \cite{spatialrobustness, translationinvariance} or varying background \cite{pupil}. Yet, most of the works in model compression focus on clean test-set accuracy ignoring model robustness, such as out-of-distribution (o.o.d.) accuracy, which is crucial for systems operating in the real world.

Model compression is also a very interesting problem from a research perspective. It is well-known that current machine learning models are heavily over-parameterized, which allows them to easily fit random labels \cite{memorization}. This over-parameterization is exploited by compression techniques which greatly reduce the model size with only a small decrease in accuracy. But, investigating only the mean accuracy might not give the full picture of compression methods’ impact on model predictions. Highly accurate models (in terms of average precision) can still fail in rare and atypical cases \cite{drbo, subclass}. 

It was only recently shown that pruning significantly affects robustness in the image classification task and might disproportionately impact different object classes \cite{hooker2019compressed, classdependent}. In our work, we start with those observations and apply them to the task of object detection from RGB images. Further, we test the effect of naturalistic data augmentation on compressed models. We focus on autonomous driving datasets, as both model robustness and computational efficiency are of great importance for such an application. It was demonstrated that using test accuracy alone might not give the complete picture of the model compression impact. Measuring out-of-distribution performance or per-class accuracy is crucial in safety-critical applications.
The contribution of this paper are as follows:
\begin{itemize}
    \item First, both structured and unstructured compression techniques are evaluated on object detection tasks. The further effect of adding texture invariant data augmentation was measured. It was shown that such an intervention has a positive effect on model robustness, showing that highly compressed models, in spite of their limited capacity, are able to build more texture-invariant object representation.
    \item When evaluating model robustness on synthetic distributional shift (adding different types of distortions to the images), it was shown that compressed models' sensitivity is remarkably varied between different distortion types and some of them are only slightly affected by the compression.
    \item It was shown that compression techniques have a disproportionate impact on different classes. To reduce that effect, several class-balancing techniques are evaluated which significantly improve accuracy for many classes, and also improve mean average precision. Noticeably, for structured pruning, the positive effect of using methods for handling data imbalance is the most striking at higher compression rates.
    
\end{itemize}

\section{Related work}

\textbf{Model compression}. A significant number of methods have been proposed for reducing the computational footprint of neural networks by reducing the model size. The most popular approach is magnitude pruning which removes a number of small magnitude weights resulting in only a small decrease in accuracy \cite{TODOnipsprune, agp}. However, in order to actually reduce the computational cost of such pruned models, specialized hardware is required which optimizes sparse operations. As a result, structured pruning was proposed where entire filters and/or layers are removed \cite{structure, todopruningfilters, cnngradual}. Standard approaches to model compression assume training the base model, pruning and then a fine-tuning stage \cite{TODOnipsprune} or gradually pruning the model during training \cite{agp, cnngradual}. For the task of visual recognition, model compression has mostly been applied to the image classification, with very few works on the task of object detection \cite{detcompr}. As object detection is a more complex task than image classification, and also has significant application potential, it is important to evaluate model compression methods in the object detection task.

\textbf{Robustness.} Current CNN-based models show impressive performance when the test data comes from a similar data distribution to the training data, but fails to generalise when this assumption does not hold \cite{recht}, which is a significant challenge when deploying machine learning models to the real world. To improve model robustness, several methods have been proposed. They include using data augmentation techniques such as style-transfer \cite{cygert, sin}, noise injection \cite{noise}, naturalistic augmentation (color distortion, noise, and blur) \cite{TODOorigins}, and interpolating between different images \cite{augmix}. It has also been shown that large models improve robustness \cite{large, Corruptions}, however our goal is orthogonal, we studied whether small models can also be robust.

On the evaluation side, evaluating the models by adding synthetic distortions during test time (e.g., noise, blur, changes in contrast) \cite{Corruptions}, using test data coming from a different distribution, and by testing models on natural transitions (for example from day to nighttime images) \cite{wintercoming} have been proposed. 
Still, measuring the effect of compression on model robustness remains an unstudied problem. It was shown that it is possible to optimise for both adversarial robustness and model size at the same time \cite{hydra, dnr}.
A parallel work to ours also investigates effects of model compression in out-of-distribution setting and confirms that such testing is critical in the context of safety-critical systems \cite{lost}.
In our work, we focus on the object detection task and measure its impact on both synthetic and real distributional shifts, and additionally on per-class accuracy.

\textbf{Class imbalance.}
Real-world datasets often follow a long-tail distribution: a few dominant classes are represented by a great number of examples, significantly higher than of other less represented classes. Models trained on such datasets provide poor accuracy on the underrepresented classes \cite{imbalance}. Significant research exists on dealing with such data imbalance which can be categorised into two groups: re-sampling and cost-sensitive learning. In re-sampling strategies, some of the training examples for the minority classes are repeated \cite{smote} or examples from dominating classes are undersampled). Cost-sensitive learning deals with the problem by assigning a relatively higher cost to the minority classes, e.g., computing the loss using the inverse of the class frequencies \cite{inverse} or the inverse of the effective 
number of samples \cite{effective}. 

At the same time, the effect of model compression techniques on certain classes remains largely unexplored. Indeed, a recent work suggests that some classes may be more impacted by compression techniques than others \cite{hooker2019compressed}. As such, we decided to evaluate the effect of model compression techniques on different classes in the safety-sensitive domain of autonomous driving.

\section{Methodology}

\subsection{Object detection}

A goal of object detection is to find where object are located in the image (object localization) and to which class they belong (object classification). Faster R-CNN \cite{Faster} is a popular algorithm in object detection that works in two stages: regions of interest selection with Region Proposal Network (RPN) is followed by a regions classification into one of the classes $c \in \{1, ..., C\}$. Both stages share a common set of convolutional layers, a so-called a backbone network. RPN outputs a list of anchors (bounding boxes) which are likely to contain an object, and each region proposal is processed by the classification layer,   which computes a logit vector $ z \in R^c$ for each region. Finally, a sigmoid function is applied $p = sigmoid(z)$, to obtain a list of predicted class probabilities and class with the highest probability is used as predicted class for given region proposal.
Whole model is trained by optimizing multi-task loss function which consists of cross-entropy loss for classification task and L1 smooth loss for bounding boxes localizations regression.

\subsection{Model compression}

In our work, standard magnitude pruning approaches were utilized. While, more advanced approaches exist, magnitude pruning has been shown to consistently achieve very good results across a number of datasets and tasks \cite{stateofsparsity}. Another advantage is that magnitude pruning is a very general method than can be applied to a wide range of tasks and architectures. During training, the automatic gradual pruning technique is used which progressively increases the sparsity in the network over the course of the training up to the desired compression rate. Specifically, the sparsity $s_t$ at epoch $t$ is computed as:\cite{agp}
\begin{equation}
    s_t = s_f + (s_i - sf) * (1 - \frac{t -t_0}{n\Delta t})^3 \; \textrm{for} \; t \in\{t_0, t_0 + \Delta t, ..., t_0 + n\Delta t \}
\end{equation}
where $n$ is the number of pruning steps, $\Delta t$ is the pruning frequency, $s_f$ is a final sparsity value, $s_i$ an initial sparsity value (usually 0) and $t_0$ is an epoch at which pruning starts.
At each iteration $L_1-$norm is computed for each tensor and tensors with the lowest norm are zeroized, such that the desired level of sparsity $s_t$ at given epoch is achieved. Similarly, for structured pruning the $L_1-$norm is computed at the filter level, which weights are set to 0.


\subsection{Data augmentation}
Several data augmentation techniques have been proposed to improve model robustness, in particular style-transfer data augmentation is quite often used \cite{sin, wintercoming, cygert}. However, the computation itself is quite costly and one has to decide which data to use as the source of the style. On the contrary, recent work has shown that adding simple data augmentation such as color distortion, noise, and blur can also be a very efficient strategy to improve model robustness \cite{TODOorigins}. As such a procedure is very simple and very efficient, it was used in our work. Namely, during training, the following augmentation is used in the pipeline:
\begin{itemize}
    \item Color distortion with a probability of 50\%. This includes changes in the brightness, contrast, saturation and hue of the image as specified in \cite{simclr}.
    \item Color drop (grayscale image) with a probability of 20\%.
    \item Gaussian blur with a probability of 50\%.
    \item Gaussian noise with a probability of 50\%.
\end{itemize}

\begin{figure}[t]
\centering
\subfloat{{\includegraphics[width=0.5\linewidth]{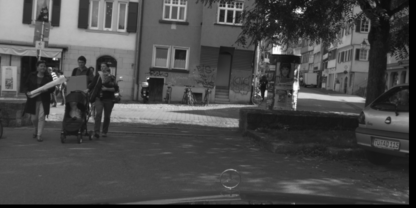} }} 
\subfloat{{\includegraphics[width=0.5\linewidth]{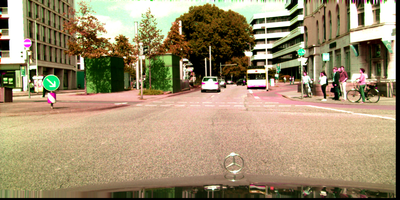} }}
\vspace{-3mm}
\subfloat{{\includegraphics[width=0.5\linewidth]{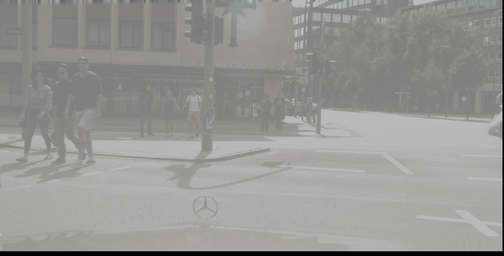} }} 
\subfloat{{\includegraphics[width=0.5\linewidth]{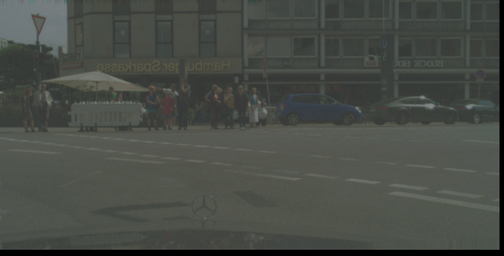} }}
\hspace{0mm}
\caption{Examples of augmented images: color drop (top left image), color distortion (top right), overexposed image (bottom left), gaussian noise (bottom right).}
\label{fig:adaptation}
\end{figure}

\subsection{Imbalanced data} \label{sec_imbalance}

In this subsection, a few techniques for handling data-imbalance are described, the first technique being based on sampling and the others based on cost-sensitive learning.

The \textbf{repeat factor sampling (RFS)} strategy was recently shown to yield competitive results on class imbalance problems \cite{lvis}. For each category $c$, let's define $f_c$ as a portion of images that contain at least one instance of object category $c$. The category-level repeat factor is defined as: 
\begin{equation}
r_c = max(1, \sqrt{(t/f_c))}
\end{equation}
where $t$ is a hyperparameter. Intuitively, this means that categories which frequency $f_c$ is below threshold $t$, will be over-sampled. Then the image-level repeat factor is computed as the maximum value over the categories in the image $i$: 
\begin{equation}
    r_i = \max_{c \in i} r_c
\end{equation}

\textbf{Cost sensitive learning}, on the other hand applies class-specific weights $w_c$ to the cross entropy loss in the classification task. For a given observation, the weighted cross entropy can be computed as:
\begin{equation}
    L_{wCE} = - \sum_{c=1}^{C} w_c * y_c log (\hat{y_c})
\end{equation}
where $y_c$ $\in \{0,1\}$ indicates whether class $c$ is the correct class for given observation and $\hat{y_c}$ is a predicted probability for class $c$, and $w_c$ is a weighting factor for every class.  
If $w_c = 1$ for all classes then the above formulation relates to the standard cross entropy loss. Below, different approaches to computing $w_c$ for data imbalance problems are briefly described.

Inverse square root of class frequency computes the $w_c$ in exactly the same way as the repeat factor $r_c$ was computed for the RFS algorithm. For our experiments, another variant was also tested where the weights were computed as $w_c = \sqrt{(t/f_c)}$ (so removing the $max$ function), which allowed the weights of some frequent classes to be smaller than $1$.   

Computing class weights by means of a category-level repeat factor, as defined above, may yield suboptimal results, since some of the images may contain just one instance of a given category, while others may contain dozens of them. As such, it was proposed in \cite{effective} to compute the weighting factors using the number of instances. Our implementation follows the details provided in \cite{overcoming}. First, the number of instances for each category $N_c$ is computed. Then, the \textbf{effective number of samples} $E_n$ for each category $c$ can be computed as:
\begin{equation}
E_{n} = \frac{1 - \beta^{Nc}}{1 - \beta}
\end{equation}
The final class weights are obtained by taking inverse of the $E_{n}$ and applying a normalisation term.

However, note that the above methods have mostly been tested on the image classification task, and object detection brings further challenges. First, object detection has a multi-task objective and scaling classification loss may introduce side effects to the overall performance (for example by changing the accuracy of the regional proposal network). Second, the above calculation does not take into account the background class (because it is hard to estimate the “frequency” of the background class, a class-weight of 1 is applied to the background class in all cases as in \cite{overcoming}). Since foreground/background separation is also a very important part of the object detection, one has to be very careful when applying different class balancing methods. As such, in the experiments section, experiments are also conducted with linearly scaled variants of the above methods.

\section{Experiments}

\subsection{Datasets}

\begin{figure}
    \centering
    \includegraphics[width=1.\linewidth]{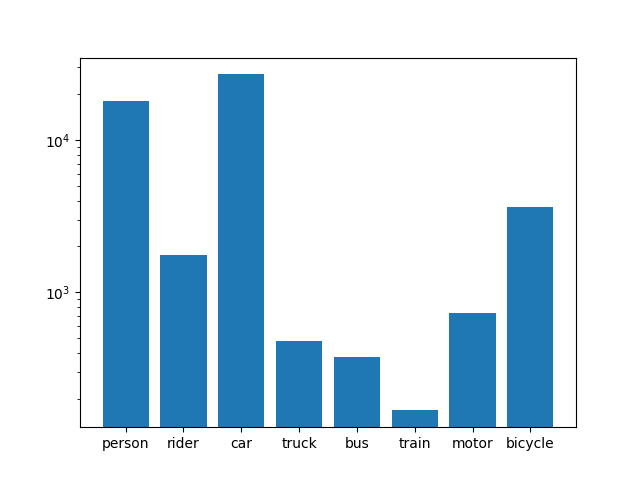}
    \caption{Cityscapes dataset class histogram (logarithmic scale).}
    \label{fig:histo}
\end{figure}

Cityscapes \cite{cityscapes} is a large-scale autonomous driving dataset for semantic segmentation and object detection (Fig. \ref{fig:histo} shows class histogram). It contains 5000 images of street scenes recorded in 27 cities, mostly in Germany. However, a potential limitation is the fact that the Cityscapes datasets were mostly recorded during the daytime in good weather conditions. As such, more challenging datasets are being developed. EuroCity Persons (ECP) \cite{Euro} contains 47,300 images recorded in 31 cities in 12 European countries. Additionally, data were recorded during all seasons in different weather conditions. A significant subset of images was recorded during the nighttime. This allows us to evaluate model robustness on a day to night transition (when the model was trained using daytime images and evaluated at nighttime). Finally, Berkeley Deep Drive (BDD) dataset \cite{bdd} was used as it is one of the most diverse datasets for object detection in autonomous driving.

However, even the biggest datasets cannot account for all different conditions that may occur in the real world, e.g., bad illumination conditions, adverse weather conditions, sensor noise, or a mixture of these. As such, using simulated distortions is often used as an additional proxy to evaluate model robustness. The common corruptions benchmark \cite{Corruptions} is a great example of such an approach, which contains procedures to generate synthetic distortions, which are applied during model evaluation. In total, 15 distortion types can be generated, which are grouped into 4 categories: noise (Gaussian noise, shot noise, impulse noise, salt-and-pepper noise), blur (defocus blur, frosted glass blur, motion blur, zoom blur), weather corruptions (snow, fog, brightness, contrast) and digital noise (elastic transformations, pixelation, JPEG lossy compression). Each corruption has 5 levels of intensity. For simplicity, in our evaluation, distortions are applied at the medium intensity level.


\subsection{Implementation details}
The models were trained using the Faster R-CNN general purpose object detector. The Distiller package \cite{distiller} was used for pruning, using both structured and unstructured methods. 
Similar to \cite{wintercoming, cygert}, the Cityscapes model was trained for 64 epochs, with a learning rate step reduction by factor of 10 at epoch 48. Initial learning rate was 0.01 and the batch size was 6 as this is the maximum that the GPU used is able to concurrently process. The pruning used the automated gradual pruning scheme \cite{agp} starting from the first epoch until epoch 56. 

For ECP and BDD datasets the model was trained for 11 epochs, with a learning rate step reduction by factor of 10 at epoch 7. Initial learning rate was 0.01 and the batch size was also 6. The pruning was gradual starting from the first epoch until epoch 8. 

The models were pruned at 30\%, 50\%, and 70\% compression rates for the structured pruning and at 50\%, 80\% and 95\% compression rates for the unstructured pruning. For each method, all of the compression rates can be considered to be a reasonable setup, with the first compression rates being more conservative and the last being more aggressive. Note, that for the unstructured pruning, higher compression rates can be achieved, which is why the compression rates were higher in that setting. Above models were trained 5 times, and the mean accuracy is reported.

\subsection{Measuring impact of model compression on the robustness}

Table \ref{tab:city1} presents the results obtained for the models trained on the Cityscapes dataset using different compression strategies and evaluated on the clean Cityscapes dataset and its corrupted versions. The results from the second to the last column measure the robustness of the models (o.o.d. test). The first thing that can be noticed is that the models clearly lack robustness and is very vulnerable to different kinds of distortions, as the mAP metric is very low for all distortion types. Further, for the structured pruning, it was possible to prune 30\% of the filters and still achieve the same accuracy on the clean dataset (first column), however models' sensitivity to different distortion types was already negatively affected. 


\begin{table}
\centering
\caption{Accuracy comparison for models trained with different pruning strategies tested on the Cityscapes dataset (first column) and different corruption types from the Common Corruptions benchmark (the remaining columns).}
\label{tab:city1}
\begin{tabular}{|l|l|l|l|l|l|}
\hline
Model & Clean & Noise & Blur & Weather & Digital\\
\hline
Baseline & 0.352 & 0.0 & 0.049 & 0.152 & 0.146\\
\hline
\multicolumn{6}{|l|}{\textit{Unstructured pruning (compression rate)}} \\
\hline
50\% & 0.351 & 0.0 & 0.047 & 0.151  & 0.14 \\
\hline
80\% & 0.338 & 0.0 & 0.041 & 0.138  & 0.135 \\
\hline
95\% & 0.323 & 0.0 & 0.029 & 0.115  & 0.118 \\
\hline
\multicolumn{6}{|l|}{\textit{Structured pruning (compression rate)}} \\
\hline
30\% & 0.352 & 0.0 & 0.037 & 0.134 & 0.135\\
\hline
50\% & 0.337 & 0.0 & 0.027 & 0.105  & 0.131 \\
\hline
70\% & 0.33 & 0.0 & 0.023 & 0.088  & 0.125 \\
\hline
\end{tabular}
\end{table}

While the previous experiment measured robustness to some synthetically generated distortions, using the ECP dataset, one can measure the robustness to natural distortion such as the transition from day to night (Table \ref{tab:ecp1}). Specifically, a model trained on daytime images is evaluated on daytime images (first column) and also on nighttime images (second column, o.o.d. test). The decrease in mAP metric is comparable for both tests, when compared to the baseline model, for both pruning methods and for both evaluation (ECP-day and ECP-night).

\begin{table}
\centering
\caption{Accuracy comparison for models trained using daytime and tested on daytime images (first column) and nighttime images (second column).}
\label{tab:ecp1}
\begin{tabular}{|l|l|l|}
\hline
Model name & ECP-day & ECP-night \\
\hline
Baseline & 0.468 & 0.392\\
\hline
\multicolumn{3}{|l|}{\textit{Unstructured pruning (compr. rate)}} \\
\hline
50\% & 0.462 & 0.396 \\
\hline
80\% & 0.45 & 0.383\\
\hline
95\% & 0.414 & 0.331\\
\hline
\multicolumn{3}{|l|}{\textit{Structured pruning (compr. rate)}} \\
\hline
30\% & 0.457 & 0.382 \\
\hline
50\% & 0.444 & 0.363 \\
\hline
70\% & 0.431 & 0.34 \\
\hline
\end{tabular}
\end{table}

\subsection{Naturalistic data augmentation}

A standard way to improve model robustness is using specialized data augmentation, it is however unclear what the effect of such augmentation will be on compressed models, especially at the highest compression rates. In this section, the models’ robustness was again evaluated, but this time a naturalistic data augmentation' \cite{TODOorigins} was used during training. The results are presented in Table \ref{tab:city2} and Table \ref{tab:ecp2}. Overall, one can see that, as expected, the out-of-distribution detection accuracy has greatly increased for both datasets for all models. For example, looking at the structurally pruned model at the 50\% compression rate, one can see that the accuracy on the distortions unseen during training has greatly increased (0.243 and 0.251 mAP for weather and digital distortions compared to 0.105 and 0.131 mAP, respectively). Also, the accuracy on the clean dataset (first column) has significantly increased for all models, with the only exception of structurally pruned model at the highest compression rate, where the accuracy has slightly decreased. Interestingly, the effect of using naturalistic data augmentation is smaller at the highest compression rates. 


On the ECP dataset, the loss in accuracy on daytime images was significant (0.15 and 0.13 at the highest compression rates for unstructured and structured pruning, however this might be because a similar decrease can be noticed for the uncompressed model (decrease in mAP from 0.468 to 0.456). This shows that one has to be careful when setting the data augmentation parameters, probably using less aggressive augmentation on the ECP dataset would improve the results for daytime images. Nevertheless, the results for the nighttime images greatly improved at all compression rates. For example, for the model structurally pruned at the 50\% compression rate, after using naturalistic data augmentation, the mAP on the nighttime images increased from 0.363 to 0.393. This shows that, in spite of limited capacity, the compressed models were still able to learn more texture-invariant representation of the objects.

\begin{table}
\centering
\caption{Accuracy comparison for models trained using naturalistic data augmentation with different pruning strategies tested on the Cityscapes dataset and corruption types from the Common Corruptions  benchmark, when using naturalistic data augmentation. Values in brackets show accuracy change due to the added augmentation.}
\label{tab:city2}
\begin{tabular}{|l|l|l|l|l|l|}
\hline
Name & Clean & Noise & Blur & Weather & Digital\\
\hline
Baseline & 0.367 (+0.015) & 0.194 & 0.126 & 0.258 & 0.271\\
\hline
\multicolumn{6}{|l|}{\textit{Unstructured pruning (compr. rate)}} \\
\hline
50\% & 0.364 (+0.013) & 0.193 & 0.127 & 0.258  & 0.264 \\
\hline
80\% & 0.359 (+0.021) & 0.18 & 0.125 & 0.255  & 0.256 \\
\hline
95\% & 0.326 (+0.003) & 0.064 & 0.112 & 0.226  & 0.233 \\
\hline
\multicolumn{6}{|l|}{\textit{Structured pruning  (compr. rate)}} \\
\hline
30\% & 0.36 (+0.008) & 0.178 & 0.122 & 0.252  & 0.252 \\
\hline
50\% & 0.352 (+0.015) & 0.154 & 0.122 & 0.243  & 0.251 \\
\hline
70\% & 0.324 (-0.006) & 0.103 & 0.113 & 0.221  & 0.232 \\
\hline
\end{tabular}
\end{table}

\begin{figure}
\centering
\includegraphics[width=0.999\linewidth]{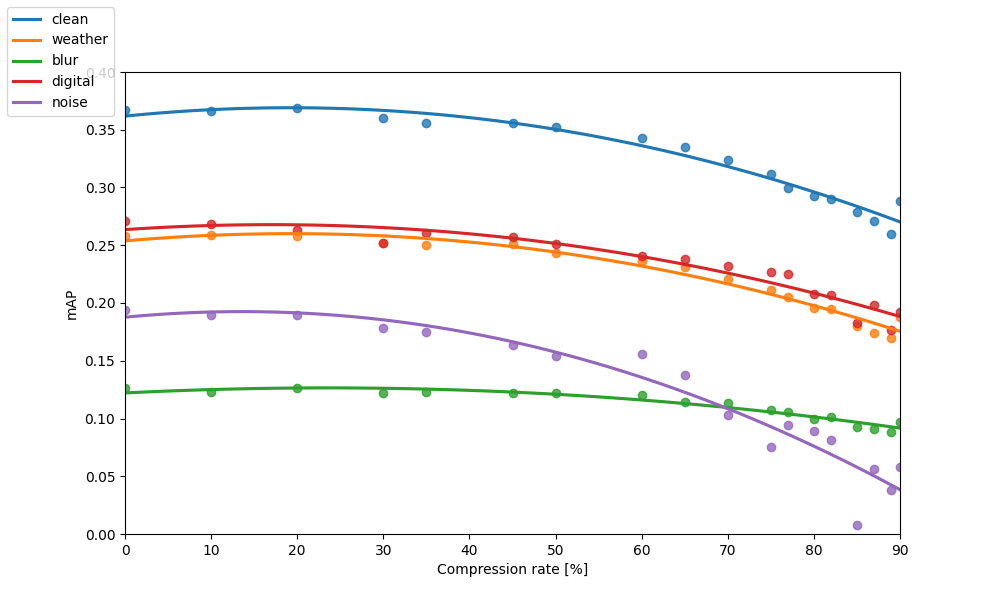}
\caption{Effect of structured pruning with different compression rates (x axis) across different distortion types on a mAP metric, for a model using naturalistic data augmentation.}
\label{fig:compare}
\end{figure}

\begin{table}
\centering
\caption{Accuracy comparison for models trained using naturalistic data augmentation on daytime images and tested on daytime images (first column) and nighttime images (second column), when using naturalistic data augmentation. Values in brackets show accuracy change due to the added augmentation.}
\label{tab:ecp2}
\begin{tabular}{|l|l|l|}
\hline
Model name & ECP-day & ECP-night  \\
\hline
Baseline & 0.456 (-0.012) & 0.419 (+0.027)\\
\hline
\multicolumn{3}{|l|}{\textit{Unstructured pruning (compr. rate)}} \\
\hline
50\% & 0.453 (-0.009) & 0.417 (+0.023)\\
\hline
80\% & 0.444 (-0.006) & 0.407 (+0.024) \\
\hline
95\% & 0.399 (-0.015) & 0.363 (+0.032) \\
\hline
\multicolumn{3}{|l|}{\textit{Structured pruning (compr. rate)}} \\
\hline
30\% & 0.447 (-0.01) & 0.407 (+0.025) \\
\hline
50\% & 0.433 (-0.011) & 0.393 (+0.03) \\
\hline
70\% & 0.418 (-0.013) & 0.381 (+0.041) \\
\hline
\end{tabular}
\end{table}
It is also worth looking at how the dynamics of change in accuracy for specific corruptions are affected, as the compression rate is increased (Fig. \ref{fig:compare}). A few, very interesting observations can be made. First, the accuracy for each corruption type was differently impacted by the pruning. The models’ sensitivity to noise was the most heavily impacted by model pruning. While the initial accuracy was fair (0.194 mAP without any compression), the accuracy started to deteriorate very quickly when more than 30\% of the filters were pruned. On the other hand, the accuracy for the blur distortions was almost flat, being only slightly reduced at the highest compression rates. Digital and weather distortions were similarly impacted by model compression, comparably to the performance of the original Cityscapes dataset. Relating the results to other work \cite{fourier}, it is worth noting that different distortions had different Fourier statistics. Some of them (i.e., shot and impulse noise) were concentrated in the high-frequency components of the image, while others (e.g., brightness, contrast) were concentrated in the low-frequency components. This might mean that pruning the visual recognition changes the models’ sensitivity to the high- and low-frequency components of the image.


\subsection{Per class evaluation}

In this section, instead of observing only the mean accuracy of the model, a per-class accuracy is also examined to see how different classes were impacted by the compression. It is important, since observing effects of compression using mean accuracy alone, may be insufficient \cite{classdependent}. In this section the experiments were conducted on Cityscapes and BDD datasets, as they provide ground-truth for many classes. It is clear that different classes were disproportionately affected by the compression techniques (Table \ref{tab:perclass}). Some classes were heavily impacted by the compression (e.g., truck, train, bus) while others were less affected (i.e, car). There are many factors which influence the final impact. One of those is the class imbalance (Fig. \ref{fig:histo}, i.e., car class is dominant in both datasets), but some classes were also inherently harder than others (because they were similar to other classes, occured with high occlusion rates, or were hard to distinguish from the background). 

As some classes seems to be more impacted than others, we have conducted experiments using methods for imbalanced datasets, namely:
\begin{itemize}
    \item Repeat factor sampling ($RFS$)
    \item Inverse squared class frequency re-weighting with ($INV_{cap}$) and without ($INV$) setting the minimal weight to be 1.0 (as described in sec. \ref{sec_imbalance})
    \item Effective number of samples ($ENS$)
\end{itemize}
For the weighting methods, we also experimented with the linear variants of the above methods using scaling factor $\lambda$~$\in~\{0.5, 1, 2.\}$ and results are reported for the best performing scale.

\begin{table*}
\centering
\caption{Per class accuracy of trained models. Aug stands for the naturalistic data augmentation and INV for the inverse class frequency re-weighting method. For the unstructured pruning, using data balancing methods bring similar gain across different compression rates, here only the accuracy at the highest compression rate is reported. }
\label{tab:perclass}
\begin{tabular}{|l|l|l|l|l|l|l|l|l|l|l|l|}
\hline
Name & Person & Rider & Car & Truck & Bus & Train & Motorcycle & Bicycle & mAP \\
\hline
\multicolumn{10}{|l|}{\textbf{Cityscapes}} \\
\hline
Baseline + aug & 0.39 & 0.404 & 0.577 & 0.265 & 0.495 & 0.222 & 0.256 & 0.329 & 0.367\\
\hline
Baseline + aug + INV & 0.39 & 0.407 & 0.576 & 0.283 & 0.512 & 0.229 & 0.263 & 0.332 & 0.374\\
\hline
\hline
Unstructured (95\%) + aug & 0.359 & 0.38 & 0.552 & 0.205 & 0.454 & 0.16 & 0.228 & 0.312 & 0.331 \\
\hline
Unstructured (95\%) + aug + INV & 0.354 & 0.378 & 0.546 & 0.221 & 0.466 & 0.186 & 0.233 & 0.31 & 0.337 \\
\hline
\hline
Structured (30\%) + aug & 0.385 & 0.409 & 0.576 & 0.249 & 0.497 & 0.186 & 0.246 & 0.333 & 0.36 \\
\hline
Structured (30\%) + aug + INV & 0.384 & 0.406 & 0.572 & 0.257 & 0.512 & 0.222 & 0.26 & 0.331 & 0.368 \\
\hline
Structured (50\%) + aug & 0.378 & 0.402 & 0.57 & 0.251 & 0.473 & 0.176 & 0.236 & 0.332 & 0.352 \\
\hline
Structured (50\%) + aug + INV & 0.379 & 0.399 & 0.567 & 0.256 & 0.502 & 0.214 & 0.245 & 0.331 & 0.362 \\
\hline
Structured (70\%) + aug & 0.362 & 0.394 & 0.559 & 0.21 & 0.432 & 0.1 & 0.221 & 0.314 & 0.324 \\
\hline
Structured (70\%) + aug + INV & 0.36 & 0.389 & 0.555 & 0.233 & 0.478 & 0.184 & 0.233 & 0.314 & 0.343 \\
\hline
\hline
\multicolumn{10}{|l|}{\textbf{BDD}} \\
\hline
Baseline + aug & 0.318 & 0.256 & 0.408 & 0.392 & 0.417 & 0.0 & 0.218 & 0.221 & 0.279\\
\hline
Baseline + aug + INV & 0.317 & 0.26 & 0.404 & 0.396 & 0.428 & 0.031 & 0.232 & 0.222 & 0.286\\
\hline
\hline
Structured (70\%) + aug & 0.266 & 0.193 & 0.388 & 0.322 & 0.339 & 0.0 & 0.150 & 0.163 & 0.228 \\
\hline
Structured (70\%) + aug + INV & 0.276 & 0.222 & 0.389 & 0.35 & 0.369 & 0.0 & 0.179 & 0.185 & 0.246 \\
\hline 
\end{tabular}
\end{table*}

Overall, very interesting results were obtained. The best performing method utilized inverse class frequency re-weighting. Interestingly, while the effect of data imbalance was relatively small without any compression (mAP increased from 0.367 to 0.374 on Cityscapes, similarly on BDD, Table \ref{tab:perclass}), the effect was much more striking at the highest compression rates, for the structurally pruned models. At the 70\% compression rate level, the accuracy significantly increased from 0.324 to 0.343 on Cityscapes and from 0.228 to 0.246 on BDD dataset. As a sanity check, models were also tested at the 75\% and 80\% compression rates, confirming those results - the overall accuracy increased by around 0.02 mAP in both cases. For the unstructured pruning, the above finding was not observed. It might occur because structured pruning is a harder problem, and in the case of model pruned with unstructured method, it might be easier to accommodate for different classes.

\begin{figure}
    \centering
    \includegraphics[width= 0.5\textwidth]{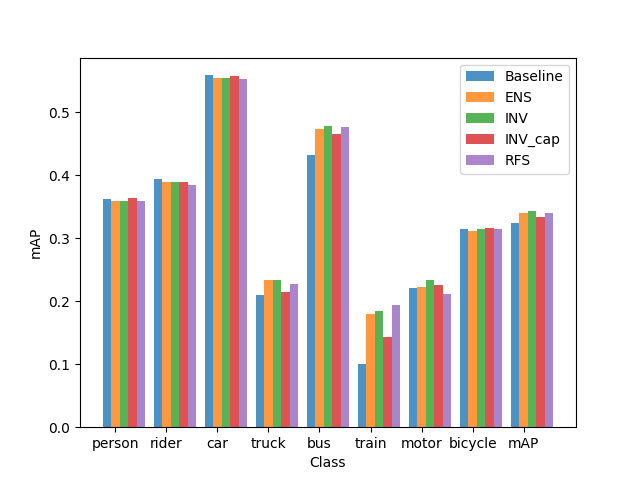}
    \caption{Per class accuracy for models structurally pruned at the 70\% compression rate using different class-balancing strategies.}
    \label{fig:barplot}
\end{figure}

Fig. \ref{fig:barplot} compares different data balancing methods on Cityscapes dataset. It can be noticed that for some classes (i.e., train, truck), the accuracy greatly increased after data balancing was applied, while on others, the accuracy remained almost the same. In general, all of the methods brought improvement to the compressed model (i.e, for INV method train accuracy increased from 0.1 to 0.184 and bus accuracy increased from 0.432 to 0.478). Recent work studied models performance of the minority groups and show that the overparameterized models seem to learn patterns that generalize well on the majority groups, but do not work well on the underrepresented classes \cite{investigation}. Our work, on the other hand studies per class accuracy on the real-world dataset in low-capacity models, and showed that different data balancing methods can be very effective (for structurally pruned models).

\newlength{\tempheight}
\newlength{\tempwidth}

\newcommand{\rowname}[1]
{\rotatebox{90}{\makebox[\tempheight][c]{\textbf{#1}}}}
\newcommand{\columnname}[1]
{\makebox[\tempwidth][c]{\textbf{#1}}}

\begin{figure*}[h!]
\setlength{\tempwidth}{.324\linewidth}
\settoheight{\tempheight}{\includegraphics[width=\tempwidth]{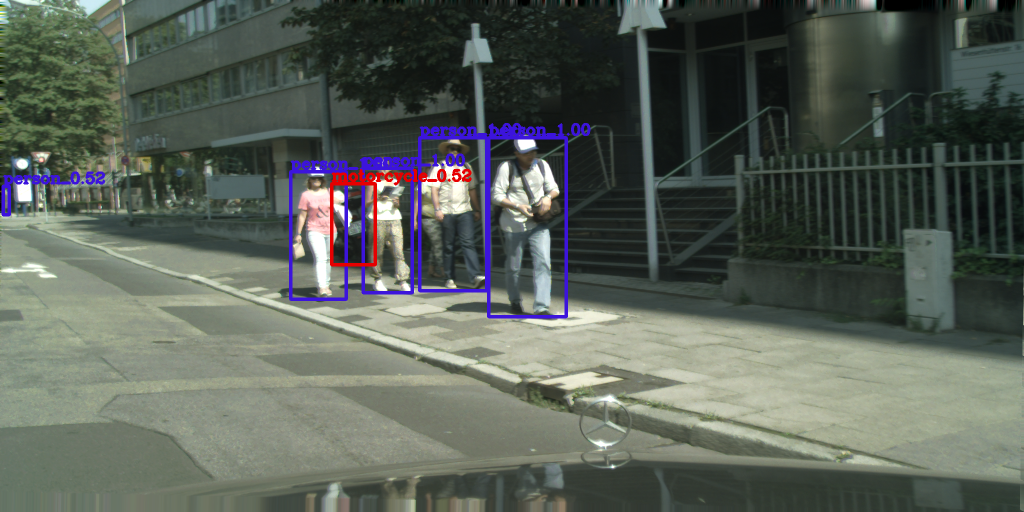}}%
\centering
\hspace{\baselineskip}
\\
\rowname{No compression}
\subfloat{\includegraphics[width=\tempwidth]{imgs/fig_5_1.png}}\hfil
\subfloat{\includegraphics[width=\tempwidth]{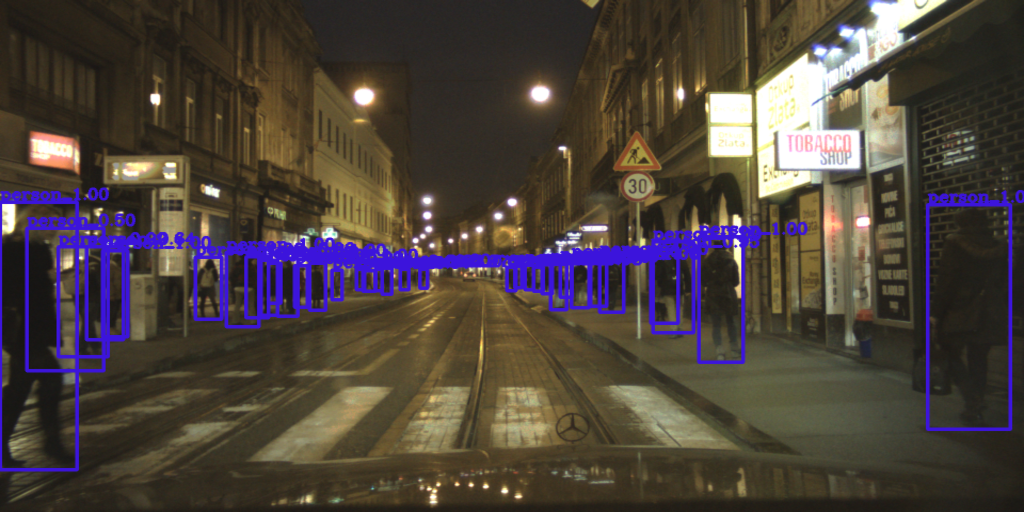}}\hfil
\subfloat{\includegraphics[width=\tempwidth]{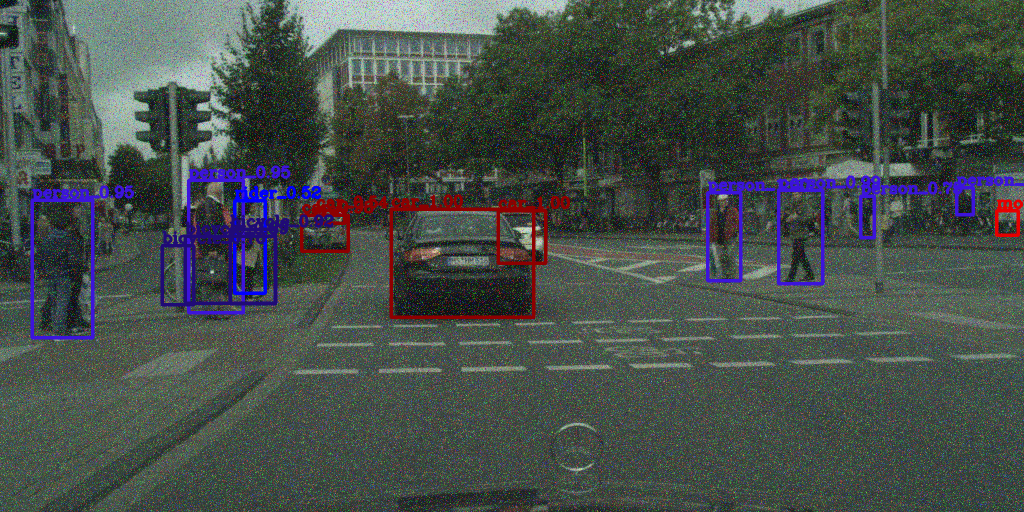}}\hfil
\rowname{Structured (70\%)}
\subfloat{\includegraphics[width=\tempwidth]{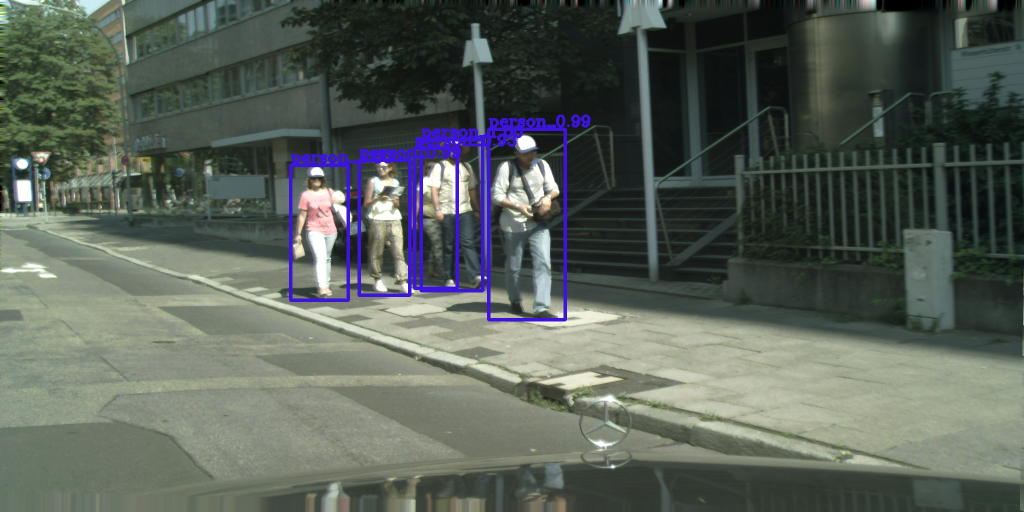}}\hfil
\subfloat{\includegraphics[width=\tempwidth]{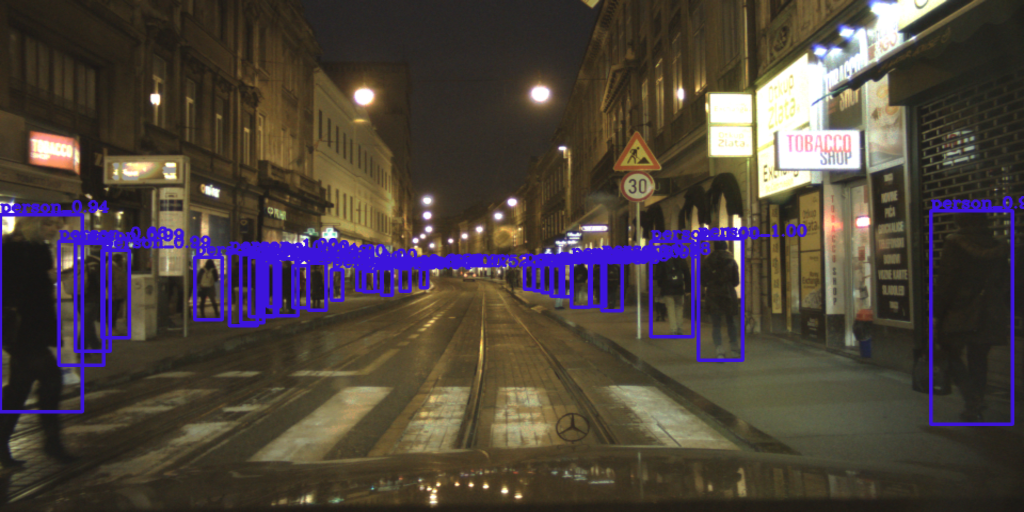}}\hfil
\subfloat{\includegraphics[width=\tempwidth]{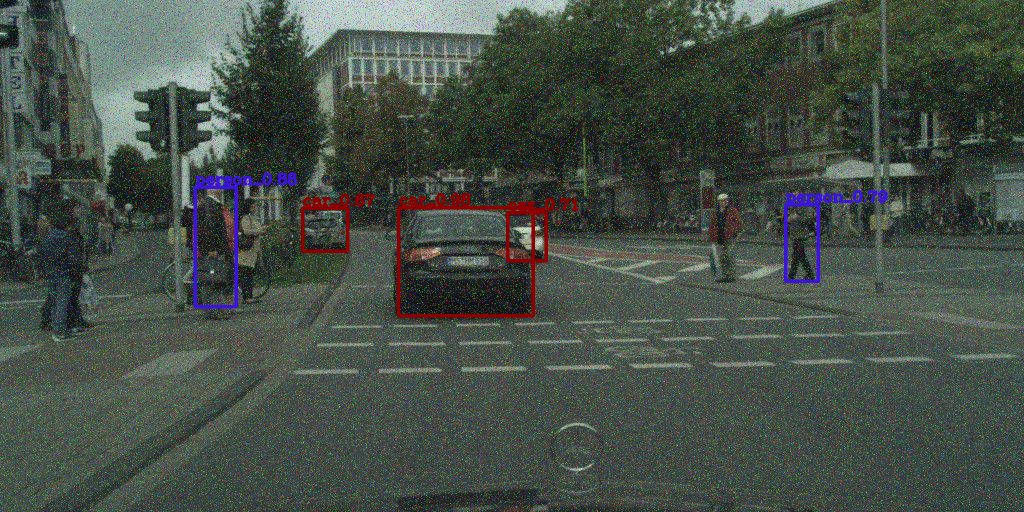}}\hfil
\caption{Detection samples for base model and the model structurally pruned at 70\% compression rate. Detections on the original Cityscapes dataset (first column), ECP dataset (second column) and noise distortion (third column). 
}
\label{fig:qual1}
\end{figure*}

Fig. \ref{fig:qual1} shows some detection examples. In general, the compressed model detects well visible objects in the image, however, the occluded objects might not be detected (the first column, missed motorcycle detection). Additionally, pruned models are much more sensitive to the noise distortion and might include more false-positive detections. 

\section{Conclusions}

In this paper, it was shown that, despite limited capacity, compressed models could make effective use of naturalistic data augmentation to learn more texture-invariant representations, which significantly increased model robustness to synthetic distortions and day to night transition. It was found that model compression differently affects models’ sensitivity to different distortion types. Some of them, i.e., those concentrated in the high-frequency domain such as Gaussian noise, were heavily affected by pruning techniques, while others (blur distortions), were only slightly affected.

In particular, it was demonstrated that data balancing methods might be especially useful in structurally pruned neural networks. 
Without any compression applied, using inverse class frequency re-weighting increased the overall mAP by 0.007 (1.9\% relative increase).
On Cityscapes dataset, at the 70\% compression rate, in the case of structured pruning, the mAP increased by 0.019 (5.9\% relative increase). Similar results were obtained for the BDD dataset. Both sampling-based methods (repeat factor sampling) and cost-sensitive methods (i.e, inverse squared class frequency re-weighting) turned out to be effective.

Overall, our work explores the relation between models' robustness and the model compression techniques and provides insights on improving both performance and computational cost of deployed models. It was shown that for safety-critical systems, testing compressed models in out-of-distribution setting or measuring per class accuracy, is important to fully understand effects of model pruning.  
A natural extension of our work would be extending our experiments with quantization techniques, which are also used for reducing computational cost of machine learning models.
As a future work, it would be also worth exploring effect of model compression on fine-grained “subclasses”, similar as in \cite{subclass}.

\bibliography{biblio.bib}
\bibliographystyle{ieeetr}

\end{document}